\documentclass{article}
\usepackage{spconf,amsmath,graphicx}
\usepackage{epsfig}
\usepackage{amssymb}

\usepackage{subfigure}
\usepackage{fancyhdr}
\usepackage{algorithm}
\usepackage{algpseudocode}
\usepackage{booktabs} % for professional tables
\usepackage{multirow}
\usepackage{xcolor}

\title{Generalized Expectation Maximization Framework for Blind Image Super Resolution}
\name{Yuxiao Li \qquad Zhiming Wang \qquad Yuan Shen}
\address{
Department of Electronic Engineering, Tsinghua University, Beijing, China  \\
Emails: \{li-yx18, wang-zm18\}@mails.tsinghua.edu.cn,
shenyuan\_ee@tsinghua.edu.cn}
\begin{document}
%\ninept
%
\maketitle
\begin{abstract}

% contain about 100 to 150 words, 12 to 18 lines
Learning-based methods for blind single image super resolution (SISR) conduct the restoration by a learned mapping between high-resolution (HR) images and their low-resolution (LR) counterparts degraded with arbitrary blur kernels. However, these methods mostly require an independent step to estimate the blur kernel, leading to error accumulation between steps. We propose an end-to-end learning framework for the blind SISR problem, which enables image restoration within a unified Bayesian framework with either full- or semi-supervision. The proposed method, namely SREMN, integrates learning techniques into the generalized expectation-maximization (GEM) algorithm and infers HR images from the maximum likelihood estimation (MLE). Extensive experiments show the superiority of the proposed method with comparison to existing work and novelty in semi-supervised learning.

\end{abstract}
\section{Introduction}
\label{sec:intro}
% 1 topic sisr
% 2 basic degradation model
Single image super resolution (SISR) refers to the restoration of the plausible and detailed high-resolution (HR) image from the corresponding low-resolution (LR) image, with a wide range of applications \cite{Siu2012ReviewOI,Dai2016IsIS}.
A widely-adopted model for SISR is as a degradation process where the LR image is a blurred, decimated, and noisy version of its HR counterpart \cite{Elad1997RestorationOA,Liu2014OnBA}. Since real-world distributions of these variables are difficult to access, restoring HR images is a classic ill-posed, inverse problem without a closed-form solution.

% 3 existing work - model-based
Learning-based methods for SISR
% based on convolutional neural networks (CNNs) 
represent to be a popular trend due to their superiority in dealing with complicated manifold-like image distributions. Pioneered by SRCNN \cite{Dong2014LearningAD}, these methods learn the mapping between LR and HR image pairs with developed neural networks, such as residual networks (ResNets) \cite{Zhang2018ResidualDN} and generative adversarial networks (GANs) \cite{Wang2018ESRGANES}. 
However, existing works typically synthesize LR images via a bicubic degradation model, and cannot generalize well to practical cases with complicated degradation settings. 

Recently, learning-based methods for blind SISR has been proposed to address such problem \cite{BellKligler2019BlindSK,Bulat2018ToLI,Gu2019BlindSW}. Most works decompose SISR into two sequential steps, and independently estimate the blur kernel from LR images before inferring the HR images \cite{Zhang2018LearningAS,Zhang2019DeepPS,Zhang2020DeepUN}. Such separation of kernel estimation and HR image restoring may not be compatible and result in error accumulation. \cite{Luo2020UnfoldingTA} proposes an end-to-end network for blind SISR where both blur kernel and HR image can be obtained simultaneously. However, the proposed method lacks a unified theoretic framework in modeling, making the estimation for real-world blur kernel remains challenging. Besides, existing techniques on the blind SISR problem are supervised and requires LR-HR image pairs, leading to a waste of unpaired LR data.

% 5 proposal: correct degradation and unified framework
We propose a unified framework based on the generalized expectation-maximization (GEM) algorithm for blind super resolution (SREMN), which can conduct the blind SISR problem in either supervised or semi-supervised manner.
Our contributions are summarized as follows:
\begin{itemize}
    \item We present a Bayesian model for image degradation.
    %, which can estimate the HR image from its LR counterpart degraded with blur kernels. 
    The model inherits a transparent interpretation and is flexible to scale to multiple types of degradations with theoretical backbone.
    \item We propose an end-to-end GEM-based network for blind SISR, 
    % which can conduct blind image restoration within a unified framework. The algorithm 
    which is applicable both in supervised schemes, and in the novel semi-supervised scheme.
   \item The proposed method shows potential capability of deep neural networks on complex problems involving latent variables by employing statistical techniques.
    \item The proposed method shows competitive results against state-of-the-art methods on blind SISR under different settings, showing the superiority in practical use.
\end{itemize}

%Though proposed for blind SISR, the inherent methodology is adaptive to a wide range of problems in computer vision. Such combination with the GEM algorithm shows great potential of introducing statistic techniques to improve the capability of advanced deep learning methods.

\section{Expectation-Maximization Framework}  %for Blind Super-Resolution
\label{sec:model}

SISR can be modeled as a degradation process where the LR image $\mathbf{y}$ is a blurred, decimated, and noisy version of an HR image $\mathbf{x}$:
\begin{equation}  \label{eq:model}
\mathbf{y}=(\mathbf{x} \otimes \mathbf{k}) \downarrow_\mathbf{s}+\mathbf{n},
\end{equation}
where $\otimes$ represents the convolution of $\mathbf{x}$ with kernel $\mathbf{k}$, $\downarrow_s$ denotes the standard $\mathbf{s}$-fold downsampler, and $\mathbf{n}$ represents environment noise commonly modeled as AWGN with given variance ${\sigma}_n^2\boldsymbol{I}$. The key challenge lies in the latent distribution over blur kernel $\mathbf{k}$, which is hard to be either pre-known or fully simulated by training data. 
% Such unknown blur kernel is one of the key challenges of blind SISR problem, leaving the restoration of HR image an ill-posed problem hard to be achieved by both reconstruction-based and learning-based solutions. 

% We model the process from a Bayesian perspective, and construct a GEM-based learning network to estimate the unknown HR image.

\subsection{Mixture blur kernel}

Suppose the blur kernel $\mathbf{k}$ can be approximated by a mixture of $L$ Gaussian kernels with a spectrum of bandwidths, where the bandwidths ${b}_l^2$ for the $l$th kernel is a random variable of an exponential distribution, i.e.,
    \begin{equation}  \label{eq: kernel}
        \begin{aligned}
        \mathbf{k}(p_x, p_y; \mathbf{b}^2) &= 1/L \sum_{l=1}^L\frac{1}{2 \pi {b}_{l}^{2}} {\exp}^{-\frac{p_x^{2}+p_y^{2}}{2 {b}_{l}^{2}}} := \mathbf{k}_{\mathbf{b}},  \\
            p(b_{l}^2) &= \mathcal{E}(b_{l}^2; \lambda_l),~l=1,\ldots,L.
        \end{aligned}
    \end{equation}
where $p_x,p_y$ are the distances from the origin in the horizontal and vertical axes, respectively.
% ${b}_l^2$ is the bandwidth of the $l$th Gaussian kernel. Each bandwidth $b_l^2$ is assumed to be a random variable of a exponential distribution with hyper-parameter $\lambda_l$. 
Denote the bandwidth vector $\mathbf{b}^2=(b_1^2,\ldots,b_L^2)^T\in\mathbb{R}^L$ and the vector of the exponential parameters $\boldsymbol{\lambda}=(\lambda_1,\ldots,\lambda_L)^T\in\mathbb{R}^L$. 
%Suppose arbitrarily given by experience (e.g. $\lambda_0=0.5$). 
% As a result, the peakiness of the Gaussian blur kernel $\mathbf{k}_b$ is randomized every time a patch is randomly cropped. 
Therefore, such blur kernel can represent the effort of a wide range of blurry types. We denote a mixture blur kernel with bandwidth vector $\mathbf{b}^2$ as $\mathbf{k}_b$.
% During learning, the distribution parameter $\lambda$ is learned by a network, which prevent the network from overfitting to a particular kernel type (comparing to learn $b^2$ itself as a vector). Instead, the network can be seen as learning a spectrum of Gaussian mixtures for kernel estimation, since in each batch, a variety of different Gaussian kernels have been used. 

\subsection{Bayesian Model}

%From the aspect of Bayesian inference, the degradation can be viewed as a generative process from the HR image $\mathbf{x}$ to LR image $\mathbf{y}$ with unknown blurry factor $\mathbf{k}_b$. % Therefore, the potential HR image $\mathbf{x}$ can be viewed as the unknown \textit{parameter} to be estimated, while the SISR problem can be transferred to the estimation of posterior distribution $p(\mathbf{x}|\mathbf{y})$. 
% Given the observed LR data for $\mathbf{y}$, the posterior of the unknown parameter $\mathbf{x}$ is hard to obtain without information from the degradation process. 
We assume that the blur kernel can be approximated by the mixture Gaussian kernel defined above.
The prior distribution for the bandwidth vector is given as:
    \begin{equation}  \label{eq:priors}
        p(\mathbf{b}^2) = \mathcal{E}(\mathbf{b}^2; \boldsymbol{\lambda}).
    \end{equation}
\noindent where $\mathcal{E}(\cdot;\boldsymbol{\lambda})$ denotes an exponential distribution with parameter $\boldsymbol{\lambda}$, arbitrarily given by experience in practice.
% (e.g. $\lambda_0=0.5$).

According to Eq.~\eqref{eq:model}, we have the likelihood distribution of the observed LR data
% conditioned on the HR image and the bandwidth vector is 
as follows:
    \begin{equation}  \label{eq:likelihood}
        p(\mathbf{y}|\mathbf{x}, \mathbf{b}^2)=\mathcal{N}(\mathbf{y}; (\mathbf{x} \otimes \mathbf{k}_{b}) \downarrow_{\mathbf{s}}, {\sigma}_n^2\boldsymbol{I}),
    \end{equation}
\noindent where $\mathcal{N}(\cdot;\mu, \sigma^2)$ denotes an Gaussian distribution with mean and variance being $\mu$ and $\sigma^2$, respectively. % The degradation settings ${\sigma}_n^2$ and $\downarrow_{\mathbf{s}}$ are pre-defined n the problem scenario.

Additionally, we assume independence between the HR image and the bandwidth vector distributions, i.e.,
\begin{equation}  \label{eq:independence}
    p(\mathbf{x},b^2)=p(\mathbf{x})p(b^2).
\end{equation}

Thus, a full Bayesian model for the problem can be obtained from Eqs.~\eqref{eq:priors}-\eqref{eq:likelihood}. 

Due to its intractability, we take the mixture kernel bandwidth $\mathbf{b}^2$ as \textit{latent data}. The \textit{complete data} of the blind SISR problem is thus $(\mathbf{y}, \mathbf{b}^2)$. The goal then turns to infer the posterior distribution of unknown \textit{parameter} $\mathbf{x}$ conditioned on the observed data $\mathbf{y}$ as well as the latent data $\mathbf{b}^2$, i.e., $p(\mathbf{x}|\mathbf{y}, \mathbf{b}^2)$.

%We formulate a statistical model for the degradation process in  Eq.\ref{eq:model} from the Bayesian perspective. Let the observed LR image $\mathbf{y}$ together with the unobserved blur kernel $\mathbf{k}$ compose the complete data of the problem. Given the mixture kernel definition in Eq.\ref{eq: kernel}, the unobserved latent data can be equivalently set to be $\mathbf{b}^2$ for convenience. Suppose the potential HR image $\mathbf{x}$ is the unknown parameter of the model to be estimated.

\subsection{GEM Algorithm}

With the complete data being $(\mathbf{y}, \mathbf{b}^2)$ and unknown parameter being $\mathbf{x}$, the unknown HR image can be obtained from the maximum likelihood estimate (MLE) of the parameter.
% by maximizing the marginal likelihood of the observed data. 
Such likelihood can be written as:
% \begin{equation}
%     \begin{aligned}
%         &\quad \log p(\mathbf{y}|\mathbf{x})  \\
%         &= \log \int_{\mathbf{b}^2} q(\mathbf{b}^2) \frac{p(\mathbf{y}, \mathbf{b}^2|\mathbf{x})}{q(\mathbf{b}^2)} d\mathbf{b}^2  \\
%         &\geq \int_{\mathbf{b}^2} q(\mathbf{b}^2) \log \frac{p(\mathbf{y}, \mathbf{b}^2|\mathbf{x})}{q(\mathbf{b}^2)} d\mathbf{b}^2  \\
%         &= \int_{\mathbf{b}^2} q(\mathbf{b}^2) \log \frac{p(\mathbf{y}|\mathbf{b}^2,\mathbf{x})p(\mathbf{b}^2|\mathbf{x})}{q(\mathbf{b}^2)} d\mathbf{b}^2  \\
%         &= \mathbb{E}_{q(\mathbf{b}^2)}\big[\log p(\mathbf{y}|\mathbf{b}^2, \mathbf{x})\big] - \operatorname{D}_{KL}\big(q(\mathbf{b}^2)\big|\big|p(\mathbf{b}^2)\big)  \\
%         &:= \mathcal{F}\big(q,\mathbf{x};\mathbf{y}\big),
%     \end{aligned}
% \end{equation}
\begin{equation}  \label{eq:bound}
    \begin{aligned}
        &\quad \log p(\mathbf{y}|\mathbf{x})  \\
        &\geq \int_{\mathbf{b}^2} q(\mathbf{b}^2) \log \frac{p(\mathbf{y}, \mathbf{b}^2|\mathbf{x})}{q(\mathbf{b}^2)} d\mathbf{b}^2  \\
        &= \mathbb{E}_{q(\mathbf{b}^2)}\big[\log p(\mathbf{y}|\mathbf{b}^2, \mathbf{x})\big] - \operatorname{D}_{KL}\big(q(\mathbf{b}^2)\big|\big|p(\mathbf{b}^2)\big)  \\
        &:= \mathcal{F}\big(q,\mathbf{x};\mathbf{y}\big),
    \end{aligned}
\end{equation}

\noindent where $\operatorname{D}_{KL}$ is the Kullback-Leibler divergence, the inequality is resulted from the Jensen's inequality and achieves equality if and only if $q(\mathbf{b}^2)=p(\mathbf{b}^2|\mathbf{y}, \mathbf{x})$. 
% (See detailed proof in Appendix.A.)

The GEM algorithm seeks to find the MLE of the marginal likelihood by iteratively applying the following two steps:

\begin{itemize}
    \item \textit{Expectation step}: $q^{(n)} = \arg \max_{q} \mathcal{F}\big(q,\mathbf{x}^{(n)};\mathbf{y}\big)$
    % \begin{equation}
    %     q^{(n)} = \arg \max_{q} \mathcal{F}\big(q,\mathbf{x}^{(n)};\mathbf{y}\big).
    % \end{equation}
    \item \textit{Maximization step}: $\mathbf{x}^{(n+1)} = \arg \max_{\mathbf{x}} \mathcal{F}\big(q^{(n)},\mathbf{x};\mathbf{y}\big)$.
    % \begin{equation}
    %     \mathbf{x}^{(n+1)} = \arg \max_{\mathbf{x}} \mathcal{F}\big(q^{(n)},\mathbf{x};\mathbf{y}\big).
    % \end{equation}
\end{itemize}

\section{Network Learning Schemes}
\label{sec:net}

In conventional cases of GEM, the two steps are conducted alternatively by optimizing the analytical expression of $\mathcal{F}\big(q,\mathbf{x};\mathbf{y}\big)$ in Eq.~\eqref{eq:bound}. However, it is hardly possible in blind SISR as the distribution of image data gets more complex and intractable. 
% Instead of making assumptions on the distributions to reduce the complexity,
We construct two neural modules to learn the optimization results for the two steps. 

% We then parameterize the intractable distributions by neural modules and conduct the optimization in learning schemes. % The proposed method can conduct blind SISR in either semi-supervised or supervised schemes, enabled by the unsupervised nature of GEM techniques.

\begin{figure}[b]
% \begin{center}
% \fbox{\rule{0pt}{2in} \rule{.9\linewidth}{0pt}}
% \end{center}
   %\vskip 0.2in
   \begin{center}
   \centerline{\includegraphics[width=0.5\textwidth]{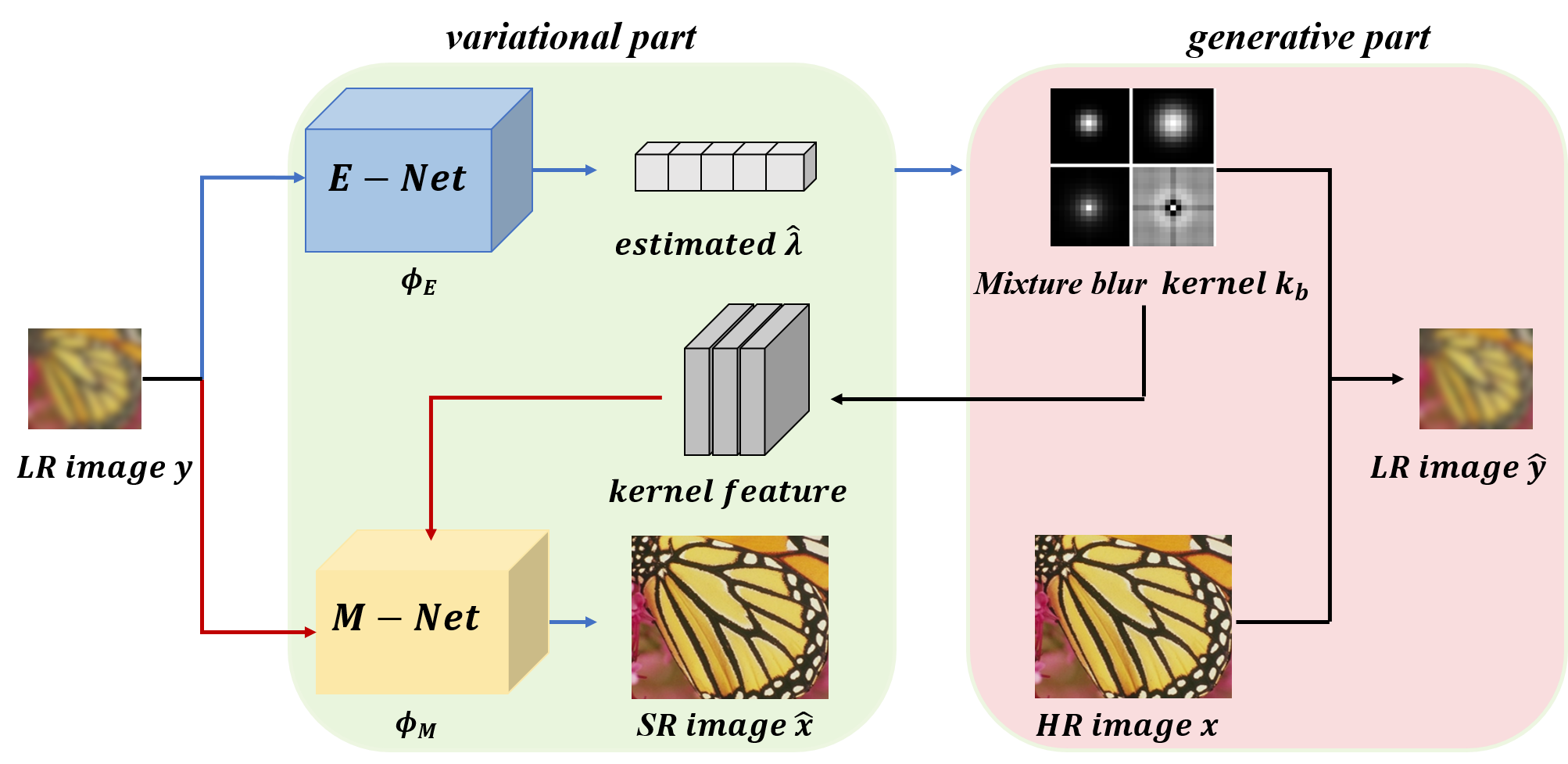}}
   \caption{The architecture of SREMN, consisting of an E-Net for blur kernel bandwidths and an M-Net for HR images.}
   %The network parameters, $\boldsymbol{\phi}_E$ and $\boldsymbol{\phi}_M$ respectively, are guided by the GEM algorithm and can be learned in either a supervised or a semi-supervised manner.}
    \label{fig:VINn}
    \end{center}
    \vskip -0.2in
\end{figure}

\subsection{Neural Modules}

% In conventional cases of GEM, the two steps are conducted alternatively by optimizing the analytical expression of $\mathcal{F}\big(q,\mathbf{x};\mathbf{y}\big)$ in Eq.~\eqref{eq:bound}. However, it is hardly possible in blind SISR as the distribution of image data gets more complex and intractable. 
% % Instead of making assumptions on the distributions to reduce the complexity,
% We construct two neural modules to learn the optimization results for the two steps. 
% In particular, we build an E-Net to learn a function $f_{\boldsymbol{\phi}_E}(\cdot)$ that maps a datapoint in the training set to a sample from the approximated posterior distribution for it, i.e., $\hat{\boldsymbol{\lambda}} = f_{\boldsymbol{\phi}_E}(\mathbf{y}) \sim q_{\boldsymbol{\phi}_E}(\mathbf{b}^2|\mathbf{y})$. Similarly, an M-Net is constructed to learn a function $g_{\boldsymbol{\phi}_M}(\cdot)$ that maps each LR sample to its estimated HR counterpart, i.e., $\hat{\mathbf{x}}=g_{\boldsymbol{\phi}_M}(\mathbf{y})$ 
%parameterized by $\boldsymbol{\phi}_E$ and a M-Net by $\boldsymbol{\phi}_M$ for the optimization of E-step and M-step, respectively.

In E-step, the optimization of $q$ is required. Following the mixture kernel assumption in Eq.~\eqref{eq: kernel}, we assume that $q$ is exponential with hyper-parameter $\hat{\boldsymbol{\lambda}}$ learned by the E-Net,
\begin{equation}
    q(\mathbf{b}^2) = \mathcal{E}(\mathbf{b}^2; \hat{\boldsymbol{\lambda}}), ~\hat{\boldsymbol{\lambda}}=f_{\boldsymbol{\phi}_E}(\mathbf{y})
\end{equation}

\noindent where $f_{\boldsymbol{\phi}_E}(\cdot): \mathbf{y}\rightarrow \boldsymbol{\lambda}$ denotes a vector-valued function parameterized by $\boldsymbol{\phi}_E$ learned with the E-Net.
%mapping from the observed data $\mathbf{y}$ to the hyper-parameter $\boldsymbol{\lambda}$ for the kernel bandwidth distribution.

In M-step, the estimation of HR image $\mathbf{x}$ is achieved by 
\begin{equation}
    \hat{\mathbf{x}}=g_{\boldsymbol{\phi}_M}(\mathbf{y}, \mathbf{k}_{\mathbf{b}})
\end{equation}
where $g_{\boldsymbol{\phi}_M}(\cdot): \mathbf{y}, \mathbf{k}_{\mathbf{b}}\rightarrow \mathbf{x}$ denotes a vector-valued function parameterized by $\boldsymbol{\phi}_M$ learned with the M-Net.
%mapping from the observed data $\mathbf{y}$ to the unknown parameter $\mathbf{x}$.

Therefore, the objective function of neural modules with respect to parameters $\boldsymbol{\phi}_E$ and $\boldsymbol{\phi}_M$ can be expressed as:
\begin{equation}  \label{eq:gem1}
    \begin{aligned}
        &\quad \bar{\mathcal{F}}(\boldsymbol{\phi}_E, \boldsymbol{\phi}_M; \mathbf{y})  \\
        % &= \mathbb{E}_{q(\mathbf{b}^2)}\big[\log p(\mathbf{y}|\mathbf{b}^2, \mathbf{x})\big] - \operatorname{D}_{KL}\big(q(\mathbf{b}^2)\big|\big|p(\mathbf{b}^2)\big)  \\
        &= \mathbb{E}_{\mathcal{E}(\mathbf{b}^2; \hat{\lambda})}\big[\log p(\mathbf{y}|\mathbf{b}^2, \hat{\mathbf{x}})\big] - \operatorname{D}_{KL}\big(\mathcal{E}(\hat{\lambda})\big|\big|\mathcal{E}(\lambda)\big)  %\\
        %&:= \bar{\mathcal{F}}(\boldsymbol{\phi}_E, \boldsymbol{\phi}_M; \mathbf{y})
    \end{aligned}
\end{equation}

\noindent where $\hat{\boldsymbol{\lambda}} = f_{\boldsymbol{\phi}_E}(\mathbf{y})$ and $\hat{\mathbf{x}} = g_{\boldsymbol{\phi}_M}(\mathbf{y},\mathbf{k}_{\mathbf{b}})$, $\mathbf{b}^2\sim\mathcal{E}(\mathbf{b}^2;\hat{\boldsymbol{\lambda}})$.

% The back-propagation (BP) algorithm for deep learning can avoid the dead-lock problem of the simultaneous optimizations, the two-step GEM can be 
Unfolded into an end-to-end learning network in Fig.~\ref{fig:VINn}, the optimization w.r.t. $\boldsymbol{\phi}_E, \boldsymbol{\phi}_M$ can be expressed as follows,

\begin{equation}  \label{eq:optimization}
    \boldsymbol{\phi}_E, \boldsymbol{\phi}_M = \arg \max_{\boldsymbol{\phi}_E, \boldsymbol{\phi}_M} \bar{\mathcal{F}}(\boldsymbol{\phi}_E, \boldsymbol{\phi}_M; \mathbf{y})
\end{equation}

\subsection{Supervised and Semi-supervised Learning Schemes}

Suppose we are given a labeled dataset $\mathcal{D}_1(\mathbf{X}, \mathbf{Y})$ with $N$ i.i.d. sample pairs $\{\mathbf{x}_i,\mathbf{y}_i\}_{i=1}^N$. Additionally we are given an unlabeled dataset $\mathcal{D}_2(\Bar{\mathbf{Y}})$, with $M$ i.i.d. LR image samples $\{\mathbf{\Bar{y}}_j\}_{j=1}^M$. Other than the common supervised learning on dataset $\mathcal{D}_1$, the proposed SREMN can utilize both data sets.

The overall loss function over the two datasets of the SREMN network is as follows:
\begin{equation}  \label{eq:semi-learning}
    \begin{aligned}
        &\quad \mathcal{L}(\boldsymbol{\phi}_E, \boldsymbol{\phi}_M; \mathbf{X},\mathbf{Y},\mathbf{\Bar{Y}})  \\
        &= \alpha_g \cdot \mathcal{L}_{GEM}(\boldsymbol{\phi}_E, \boldsymbol{\phi}_M; \mathbf{Y},\mathbf{\Bar{Y}}) + \alpha_r \cdot \mathcal{L}_{REG}(\boldsymbol{\phi}_E; \mathbf{X},\mathbf{Y}),
    \end{aligned}
\end{equation}

\noindent where $\alpha_g, \alpha_r$ are fine-tuning weights to balance the effect of the GEM term and the regularization term over the given datasets. It can be seen that the amount of supervision can be safely adjusted by the number of samples in the two datasets.

The first term $\mathcal{L}_{GEM}(\boldsymbol{\phi}_E, \boldsymbol{\phi}_M; \mathbf{Y},\mathbf{\Bar{Y}})$ is unsupervised, set to be the negative expectation of the GEM objective function over the given datasets, expressed as:
\begin{equation}
    \mathcal{L}_{GEM}(\boldsymbol{\phi}_E, \boldsymbol{\phi}_M;\mathbf{Y}, \mathbf{\bar{Y}}) = - \mathbb{E}_{\mathbf{y}\sim \mathbf{Y},\mathbf{\Bar{Y}}}\big[\bar{\mathcal{F}}(\boldsymbol{\phi}_E, \boldsymbol{\phi}_M;\mathbf{y}) \big].
\end{equation}

The second term is a supervised one on $\mathcal{D}_1$ for regularization, 
% set to be the expected mean square between the ground-truth and the estimated HR images by the M-Net, 
expressed as:
\begin{equation}
    \mathcal{L}_{SUP}(\boldsymbol{\phi}_M; \mathbf{X}, \mathbf{Y}) = \mathbb{E}_{\mathbf{x},\mathbf{y}\sim \mathbf{X}, \mathbf{Y}}\big[\Vert \mathbf{x} - g_{\boldsymbol{\phi}_M}(\mathbf{y}) \Vert^2 \big].
\end{equation}

Note that if there exist ground-truth kernels in the training set, i.e., $\mathcal{D}_k(\mathbf{K})=\{\mathbf{k}_i\}_{i=1}^N$ paired with $\mathcal{D}_1(\mathbf{X}, \mathbf{Y})=\{\mathbf{x}_i, \mathbf{y}_i\}_{i=1}^N$ is given, the supervised term can include the constraint on estimated kernels, expressed as:
\begin{equation}
    \begin{aligned}
        \mathcal{L}_{SUP}(\boldsymbol{\phi}_M; \mathbf{X}, \mathbf{Y}, \mathbf{K})
        =& \mathbb{E}_{\mathbf{x},\mathbf{y}\sim \mathbf{X}, \mathbf{Y}}\big[\Vert \mathbf{x} - g_{\boldsymbol{\phi}_M}(\mathbf{y})\Vert^2\big]  \\
        &+ \mathbb{E}_{\mathbf{k}\sim\mathbf{K}}\big[\Vert \mathbf{k} - \mathbf{k}_{\mathbf{b}}\Vert^2\big].
    \end{aligned}
\end{equation}

% \noindent where $\mathbf{k}_{\mathbf{b}}$ is the estimated kernel formed from the estimated bandwidth $\mathbf{b}^2$ sampled from the exponential distribution with parameter $\hat{\boldsymbol{\lambda}}=f_{\boldsymbol{\phi}_E}(\mathbf{y})$, as claimed above. In practice such kernels can be reduced by feature extraction techniques (e.g. PCA methods) before calculating the constraint. 

% It can be seen that the amount of supervision can be safely adjusted by the number of samples in the two datasets, i.e. $M$ and $N$. 

% \begin{itemize}
%     \item If $M\neq 0, N=0$, the SREMN conducts blind SISR in a fully unsupervised way. 
%     \item If $M, N\neq 0$ with some suitable ratio, the SREMN conducts blind SISR in a semi-supervised way. 
%     \item A fully supervised method can also be obtained if $M=0$, i.e., only dataset $\mathcal{D}_1$ with labeled data is given. 
% \end{itemize}

% Note that $\alpha_g\neq 0$ is always the case in our setting. If $\alpha_g$ is set to zero, the formulation falls to the common cases of SISR methods and cannot perform blind SISR with arbitrary kernels.

\section{Experiments}

\begin{table*}[htbp]
\caption{Quantitative comparisons with SOTA SR methods. Best results are highlighted in red and blue respectively.}
\label{tab:set2}
%\vskip 0.15in
\begin{center}
\begin{small}
\begin{sc}
\scriptsize
\centering
\setlength{\tabcolsep}{6.5mm}{
\begin{tabular}{{|c|c|c c|c c|}}
\hline
%\multirow{2}{*}
{Types} & %\multirow{2}{*}
{Method} & \multicolumn{2}{c|}{Scale $\times 2$} & \multicolumn{2}{c|}{Scale $\times 4$}\\\
% \cline{3-6} \cline{7-10} \cline{11-14}
	& & PSNR & SSIM  & PSNR & SSIM  \\
\hline
\hline
%\multirow{3}*
{Class $1$}
	& Bicubic
     & 28.73     & 0.8040    & 25.33     & 0.6795  \\
	& Bicubic kernel + ZSSR \cite{Shocher2018ZeroShotSU}
    & 29.10     & 0.8215    & 25.61     & 0.6911  \\
    & EDSR \cite{Lim2017EnhancedDR}
    & 29.17     & 0.8216    & 25.64     & 0.6928  \\
    & RCAN \cite{Luo2020UnfoldingTA}
    & 29.20     & 0.8223    & 25.66     & 0.6936\\
\hline
\hline
%\multirow{5}*
{Class $2$}
	& PDN \cite{Ma2020ImageSV} - $1$st in NTIRE'19 track4
    & /   & /     & 26.34     & 0.7190    \\
	& WDSR \cite{Yu2018WideAF}- $1$st in NTIRE'19 track2
    & /   & /     & 21.55     & 0.6841    \\
    & WDSR \cite{Yu2018WideAF}- $1$st in NTIRE'19 track3
    & /   & /      & 21.54     & 0.7016    \\
    & WDSR \cite{Yu2018WideAF}- $2$nd in NTIRE'19 track4
    &  /    & /       & 25.64     & 0.7144    \\ 
    & Ji \textit{et al.} \cite{Ji2020RealWorldSV} - $1$st in NTIRE'20 track1
    &  /  & /     & 25.43     & 0.6907 \\
\hline
\hline
%\multirow{7}*
{Class $3$}
	& Cornillere \textit{et al.} \cite{Cornillre2019BlindIS}
    & 29.46     & 0.8474    & /     & /   \\
	& Michaeli \textit{et al.} \cite{Michaeli2013NonparametricBS} + SRMD \cite{Zhang2018LearningAS}
    & 25.51     & 0.8083    & 23.34     & 0.6530    \\
    & Michaeli \textit{et al.} \cite{Michaeli2013NonparametricBS} + ZSSR \cite{Shocher2018ZeroShotSU}
   & 29.37     & 0.8370    & 26.09     & 0.7138    \\
    & KernelGAN \cite{BellKligler2019BlindSK} + SRMD \cite{Zhang2018LearningAS}
    & 29.57     & 0.8564    & 25.71     & 0.7265    \\
    & KernelGAN \cite{BellKligler2019BlindSK} + USRNet \cite{Zhang2020DeepUN}
    & /          &   /             & 20.06     & 0.5359 \\
    & KernelGAN \cite{BellKligler2019BlindSK} + ZSSR \cite{Shocher2018ZeroShotSU}
    &30.36
    &0.8669
    &26.81
    &0.7316 \\
\hline
\hline
%\multirow{2}*
{Class $4$}   
    & DAN \cite{Luo2020UnfoldingTA}
    &\textcolor{red}{32.56} &
    \textcolor{blue}{0.8997}  &\textcolor{blue}{27.55} &\textcolor{blue}{0.7582}  \\
    & SREMN (OURS)
    & \textcolor{blue}{32.25} & \textcolor{red}{0.9370} & \textcolor{red}{27.74} & \textcolor{red}{0.8015}  \\
\hline
\end{tabular}
}
%\caption{Benchmark results of state-of-the-art SR methods which were trained in 291 images: Average PSNR/SSIM/IFC for $\times$2, $\times$3, and $\times$4 upscaling. The emphasized figures indicate the best performance.}
\end{sc}
\end{small}
\end{center}
%\vskip -0.1in
\end{table*}

\subsection{Experimental Setup}

% \subsubsection{Data for Training and Testing.} 
We adopt the experimental setting for training and testing first introduced in \cite{BellKligler2019BlindSK} and utilized in \cite{Luo2020UnfoldingTA}. The training set includes $3450$ HR images from DIV2K \cite{Agustsson2017NTIRE2C} and Flick2K \cite{Timofte2017NTIRE2C}, with LR images synthesized by anisotropic Gaussian kernels. Benchmark dataset DIV2KRK is utilized as the testing set.
%We also conduct experiments on real world images, \eg \emph{chip} \cite{Fattal2007ImageUV},  to prove the generalization of SREMN in practical use.

% \subsubsection{Hyperparameters. }
The Adam \cite{Kingma2015AdamAM} optimizer is adopted for training. The learning rate is $0.0002$, and the decay of first and second momentum of gradients are $\beta_1=0.9$, $\beta_2=0.99$, respectively. All models are trained for $1000$ epochs with batch size $16$ on a single RTX1080Ti GPU.

We apply PSNR and SSIM metircs for image quality assessment, calculated on the Y-channel of transformed YCbCr space.
%, with removal of a \emph{s}-pixel wide strip from each $\times$\emph{s} upscaling image border.
We compare our method with competitors from four different classes four completeness, shown in Table~\ref{tab:set2}.

% \begin{itemize}
%     \item \emph{Class 1}: SOTA non-blind SR methods trained on bicubically downsampled images, including ZSSR \cite{Shocher2018ZeroShotSU} and RCAN \cite{Zhang2018ImageSU}.
    
%     \item \emph{Class 2}: Blind SR methods designed for NTIRE'2018 Blind-SR challenge \cite{Timofte2018NTIRE2C}, including PDN \cite{Ma2020ImageSV} and WDSR \cite{Yu2018WideAF}.
    
%     \item \emph{Class 3}: Two-step solutions to blind SR, consisting of a kernel estimation method and a separated non-blind SR method, such as the combination of Kernel-GAN \cite{BellKligler2019BlindSK} and ZSSR \cite{Shocher2018ZeroShotSU}.
    
%     \item \emph{Class 4}: Blind SR method within a single step, including DAN \cite{Luo2020UnfoldingTA} and the proposed SREMN.
% \end{itemize}

\begin{figure}[htbp]
% \begin{center}
% \fbox{\rule{0pt}{2in} \rule{.9\linewidth}{0pt}}
% \end{center}
   %\vskip 0.2in
   \begin{center}
   \centerline{\includegraphics[width=0.45\textwidth]{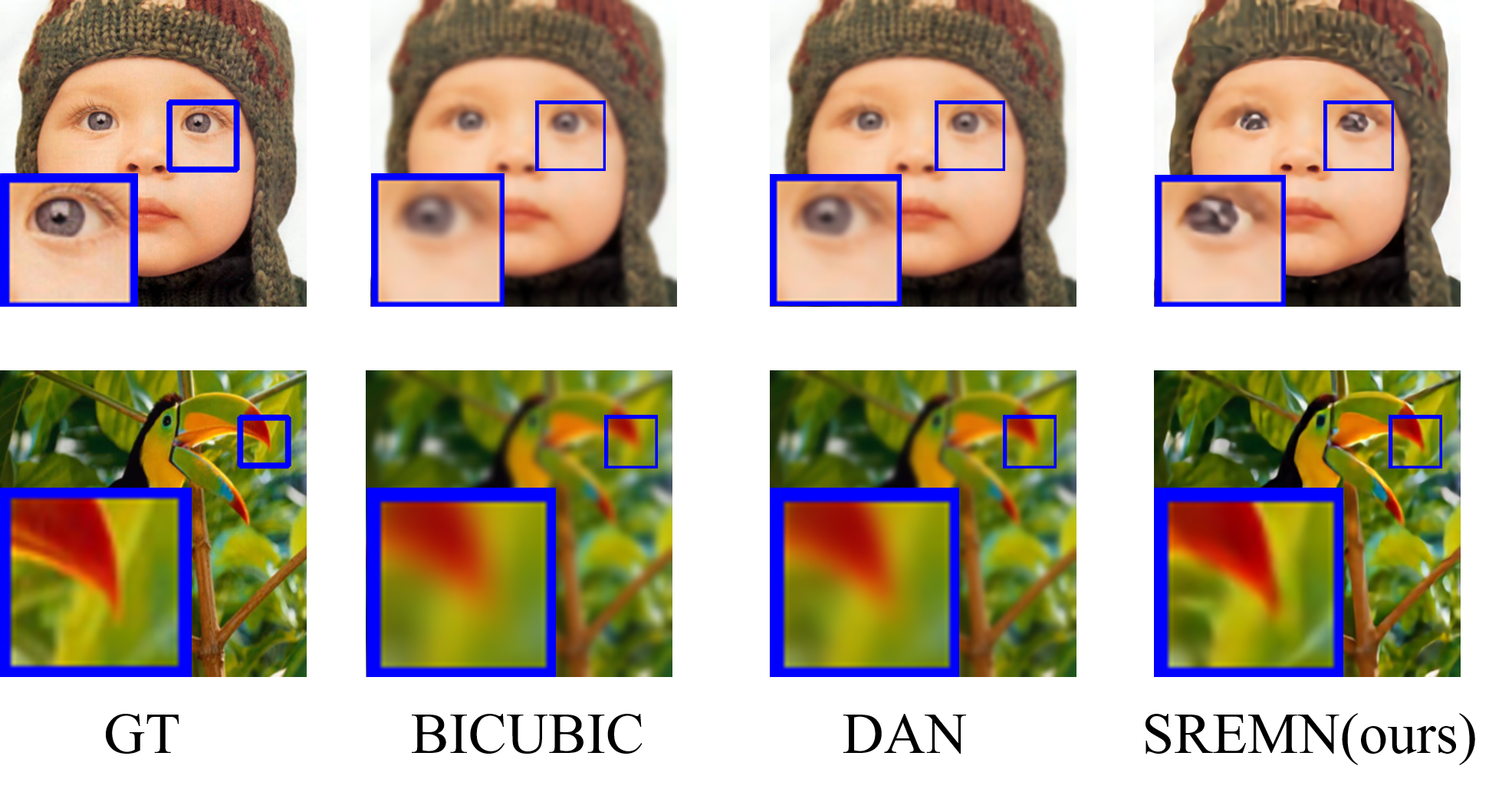}}
   \caption{Results of \emph{img} $001$ and $002$ in Set5 (bandwidth 3.2).}
    \label{fig:s1}
    \end{center}
    \vskip -0.2in
\end{figure}

\begin{figure}[htbp]
% \begin{center}
% \fbox{\rule{0pt}{2in} \rule{.9\linewidth}{0pt}}
% \end{center}
   %\vskip 0.2in
   \begin{center}
   \centerline{\includegraphics[width=0.5\textwidth]{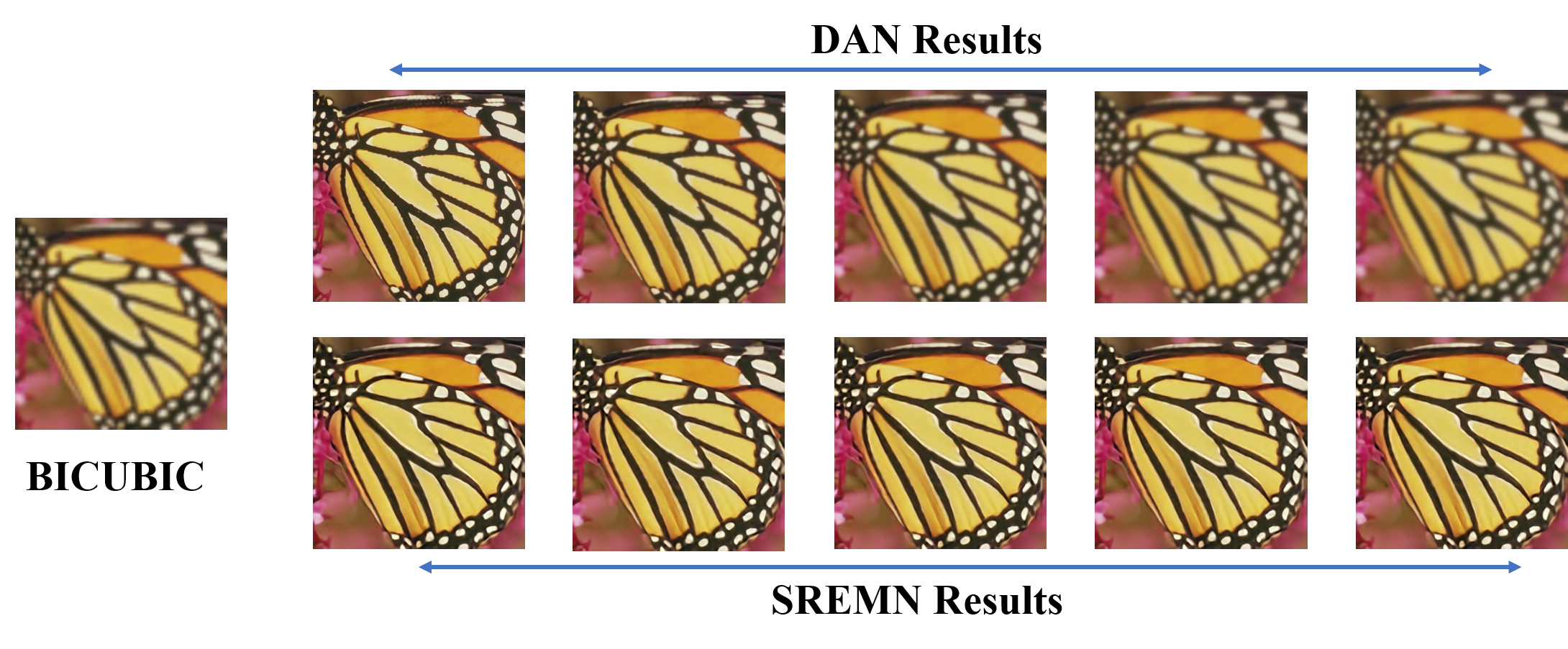}}
   \caption{Results of \emph{img} $003$ in Set5 with the kernel bandwidths in $[1.8, 2.2, 2.6, 3.0, 3.2]$. It can be seen that the proposed SREMN is more robust over different settings.}
    \label{fig:s2}
    \end{center}
    \vskip -0.4in
\end{figure}

\begin{figure}[!b]
    \centering
    \centerline{\includegraphics[width=0.45\textwidth]{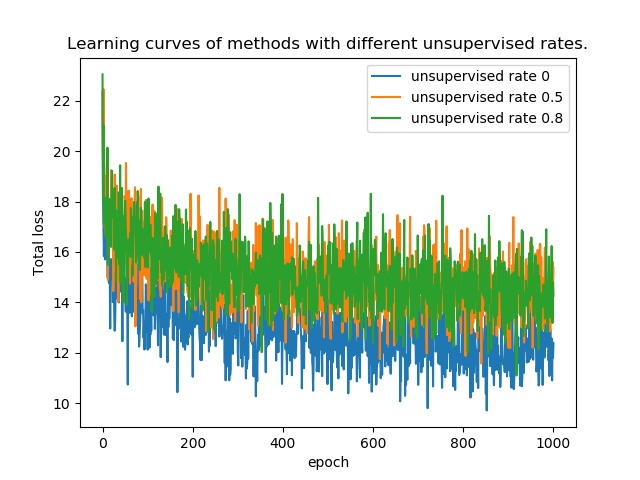}}
    \caption{Learning curves under different unsupervised rates. A higher rate results in a more challenging convergence of training.}
    \label{fig:semi} 
\end{figure}

\subsection{Quantitative Results}

Our method outperforms SOTA SR results of Blind SR with sequential steps while achieves similar results as DAN \cite{Luo2020UnfoldingTA}, a single step Blind SR method. Quantitative results are shown in Table~\ref{tab:set2}. Our method gets nice ranks in both evaluation metrics, as it encourages diverse outputs with semantic variables and also has a well-defined objective function to regularize the generation. Though a little inferior to DAN in some cases, the proposed method shows some improvements in the qualitative perspective, as illustrated in Fig.~\ref{fig:s1}.
The inferiority may result from the following reasons: 1) We use a single RTX1080Ti GPU for training, while authors of DAN use distributed training with $8$ RTX2080Ti GPUs. %Distributed training and the more updated version of hardwares may contribute to increments in their performance. 
2) Our model is trained with the batch size of $16$, while DAN is trained with the batch size of $64$, enabled by distributed training. 
%The larger version over data samples may contribute to a faster convergence and better optimal for learning algorithms. 
%3) Our method appears to be more robust to kernel variants, as illustrated in Fig.\ref{fig:s2}. The novel semi-supervised learning scheme also suggests the robustness of our method, since it requires less labeled data resources. While robustness and performance always compose a trade-off for algorithm developing, the proposed algorithm is promising to offer insights of improvement to some extent.
Therefore, the proposed method is considered to have comparable and more robust performance compared to DAN.

\begin{table}[t]
\caption{Quantitative results of the proposed method with different unsupervised rates.}
\label{tab:sup}
%\vskip 0.15in
\begin{center}
\begin{small}
\begin{sc}
\begin{tabular}{l|cccr}
\toprule
Methods & PSNR & SSIM & Time (H) \\
% \midrule
% BICUBIC & 0.1244 & 0.1244  \\
\midrule
\textit{Semi-SREMN ($\eta=0$)}   & 30.4240 & 0.9252 & 40.7  \\
\textit{Semi-SREMN ($\eta=0.5$)}   & 30.4901 & 0.9262 & 57.5  \\
\textit{Semi-SREMN ($\eta=0.8$)}   & \textbf{30.7218} & \textbf{0.9300} & 82.6  \\
\bottomrule
\end{tabular}
\end{sc}
\end{small}
\end{center}
\vskip -0.2in
\end{table}

\subsection{Semi-supervised Learning}

Additional to $N$ samples of LR-HR pairs claimed above, we introduce $M$ LR samples without according HR samples to the training to see if our method can utilize such information. We define the unsupervised rate of the network learning as $\eta=M/(M+N)$, which increases with the amount of additional unlabeled data.
% , where $M$ is the number of paired LR-HR samples and $N$ is the number of LR samples without according HR samples provided, as claimed in sec.$4.3$. % The PSNR performance of with supervision rate ranging from $0.2$ to $1.0$ is illustrated in Fig.~\ref{fig:semi}.
We compare performances of the proposed approach under $3$ different supervision rates, i.e., $\eta=0, 0.5, 0.8$ for $M=0, N, 4N$.

Quantitative results are shown in Table~\ref{tab:sup}. It can be seen that models trained with higher supervision rate tend to have better results. This validate the proposed methodology that the potential of unlabeled data could be excavated by probabilistic modeling.

The learning curves of the proposed SREMN under different supervision rates are illustrated in Fig.\ref{fig:semi}. It can be seen that the learning processes converges at around $600$ epochs, while the methods with higher supervision rates converge slower though with better ultimate performances, in accordance with intuition.

\section{Conclusion}

We propose a GEM framework for blind SISR (SREMN), which embeds the general image degradation model in a Bayesian model and enables efficient estimation of HR images.
% with GEM algorithm implemented by deep neural networks. 
The proposed method leverages benefits from both model-based methods and learning-based methods, novel in conducting the blind SISR in a semi-supervised manner. 
% On one hand, the presented method interprets the HR images as the unknown parameters estimated by MLE, leading to a sound theoretical backbone from statistics and improved scalability towards different degradation scenarios. On the other hand, the method utilizes deep neural networks to learn the approximated distributions and parameters in the GEM algorithm, which greatly integrates the efficiency of deep learning techniques. 
% Although proposed for blind SISR, it provides a promising methodology for embedding deep learning in Bayesian modeling, potential to benefit a wide range of learning problems involving a complicated process with latent variables. 
Future work would be focused on a more flexible framework on image enhancing, integrating multiple related tasks.

\vfill\pagebreak

% -------------------------------------------------------------------------
\bibliographystyle{IEEEtran}
\bibliography{refs}

% \bibliographystyle{ieee_fullname}
% \bibliography{egbib}

\end{document}